\documentclass[review,authoryear,10pt]{elsarticle}

\usepackage{latexsym}
\usepackage{amssymb}
\usepackage[T1]{fontenc}
\usepackage{booktabs}
\usepackage{amsmath}
\usepackage{graphicx}
\usepackage{amsmath}
\usepackage{xcolor,float}
\usepackage[english,dutch]{babel}
\usepackage{soul}
\usepackage[left=1in,top=1in,right=1in,bottom=1in,nohead]{geometry}

\journal{arXiv}

\begin{document}
	
	\begin{frontmatter}
		\title{Image-Based Sorghum Head Counting When You Only Look Once}
		
		\begin{abstract}
    Modern trends in digital agriculture have seen a shift towards artificial intelligence for crop quality assessment and yield estimation. In this work, we document how a parameter tuned single-shot object detection algorithm can be used to identify and count sorghum heads from aerial drone images. Our approach involves a novel exploratory analysis that identified key structural elements of the sorghum images and motivated the selection of parameter-tuned anchor boxes that contributed significantly to performance. These insights led to the development of a deep learning model that outperformed the baseline model and achieved an out-of-sample mean average precision of 0.95.
		\end{abstract}
		
		\begin{keyword}
			Deep learning, computer vision, object detection, YOLO, agriculture
		\end{keyword}

		\author[oag]{L.~Mosley}	
		
		\author[imse]{H.~Pham\corref{cor1}}
		\ead{htpham@iastate.edu}	
		
		\author[oag]{Y.~Bansal}
		
		\author[oag]{E.~Hare}

		\cortext[cor1]{Corresponding author}
	    \address[oag]{Omni Analytics Group, Ames, Iowa USA}
		\address[imse]{Department of Industrial and Manufacturing Systems Engineering, Iowa State University, Ames, Iowa USA}

	\end{frontmatter}

\section{Introduction}\label{sec1}

Phenotyping and genotyping comprise two main areas of plant breeding. Phenotyping involves the measurement of an observable trait, whereas genotyping studies the genetic composition of plants. While recent technological advancements have made genotyping more accurate, faster, and affordable, phenotyping has become the bottleneck in accelerating breeding programs (\cite{Tester818}). Certain phenotypic traits, such as yield, are accurately and efficiently measured during the harvest process. However, other phenotypes, such as stalk strength and head count, require labor-intensive, expensive, and error prone manual intervention. The advent of modern technology combined with the need to innovate has created a cultural shift towards digital agriculture. The use of self-driving equipment, as well as drone imagery and object recognition software, provide a glimpse towards the future of agriculture (\cite{Cariou,Tripicchio,thomas2021intelligent}). 

Image-based algorithms for the detection and counting of crops have been applied to corn, grapes, tomatoes, apples, and mangoes, but these approaches typically require high-resolution images captured with minimum signal to noise ratios (\cite{Qureshi2017,Gnadinger2017,khaki2022wheatnet}). With advancements in unmanned aerial vehicles, drone imagery, and machine learning, we are able to progress towards a digital future where labor-intensive phenotyping is no longer required (\cite{shaikh2022towards}). For the future of agriculture, merging these components will allow for a low cost way to monitor or estimate crop yield to identify low performing areas and detect damaged crops. For a farmer who manages more than 10,000 acres of land, traversing each field is not a feasible option.  The ability for captured aerial field images to alert farmers of any in-field phenotypic variation is vital for making real-time decisions on managing fields.

\begin{figure}[!h]
	\centering
	\includegraphics[width=0.26\linewidth,angle=90]{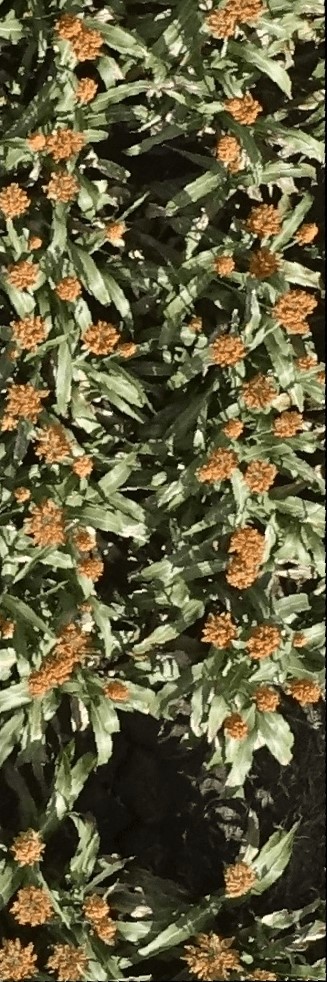}
	\caption{Image of sorghum grass taken with a UAV. The individual heads are the circular mild yellow-ish objects resting on top of the green curvi-linear leaves.}
	\label{fig:sorgsingleimage}
\end{figure}

In the case of sorghum (\textit{Sorghum bicolor}), being able to monitor the growth stages and color of their heads informs the farmer of the health and quality of their crop. With this information, farmers are able to manage decisions on how to maximize the growth potential of their sorghum by possibly spraying pesticides, adding fertilizers, etc. Current practice leans on manual, labor intensive counting of heads in a field.  For small to medium sized operations, this may not present a significant burden. However, in commercial breeding programs and traditional farms, the number of sorghum plots often exceeds thousands of acres. The sheer number of plots makes it infeasible to accurately count all heads, limiting the effectiveness of manual approaches.  This manual labor bottleneck is one of many motivating factors for combining drone imagery and machine learning.  Taken by an autonomous drone, Figure \ref{fig:sorgsingleimage} is an example overhead shot of two rows of sorghum crops.  %With images of this form, \textcite{Guo2018, Ghosal2019} both applied machine learning techniques to count sorghum heads, one uses a quadratic support vector machine while the other utilizes a deep learning approach called RetinaNet.  Analyzing drone images of sorghum fields, both were able to accurately count sorghum heads. These two approaches demonstrate the potential that machine learning can play in automatizing labor intensive agricultural tasks. 
Instead of traversing hundreds of acres of land to gather data and crop characteristics manually, commercial breeding programs and traditional farmers can utilize  advancements in technology to enable real-time decision making.

%Not only did these papers pave the way forward in demonstrating the potential drones have with regards to digital agriculture, but they also show a glimpse into what is possible for the future of agriculture.

Specifically, for this paper,
\begin{itemize}
    \item we aim to provide an approach to count sorghum heads by way of the single-shot ``You Only Look Once'' object detection algorithm (\cite{bochkovskiy2020yolov4}), and
    \item we provide insights into the data through that help motivate the tuning of YOLO parameters through detailed exploratory data analysis.
\end{itemize}
For simplicity, we will be referencing the ``You Only Look Once'' Version 4, YOLOv4, implementation for this work, and for simplicity, will simply refer to this as YOLO for the duration of this paper. We demonstrate that accurate detection results for plant phenotyping, in particular sorghum head detection, are obtainable through anchor box tuning and can thus mitigate the bottleneck imposed by manual counting methods. Additionally, we provide insights into the data through that help motivate the tuning of YOLO parameters. To achieve this goal, section 2 provides an exploratory analysis on the sorghum image. Section 3 outlines the methodology of our approach, while results. Lastly, the discussion is elaborated in Section 4 before concluding with Section 5.

\section{Data and Exploratory Methods}\label{sec2}

350 aerial images of sorghum heads were used for this analysis, separated into 300 labeled training and 50 unlabeled test images. Structurally, each training instance is a duple containing both an image and structured text file containing the locations, in pixels coordinates, of the individual bounding boxes for each identified sorghum head in the image. This structure, found in Table \ref{tab:training}, defines the class and the four edge coordinate locations of the box containing the plant head. These coordinates can then be linked back to the source image, such as the one seen in Figure \ref{fig:sorgsingleimage}

\begin{table}[h!]
	\centering
	\caption{Sample bounding box coordinates for sorghum heads in the training set.}	
	\label{tab:training}
	\begin{tabular}{lllll}
			\toprule
			Class Name            & Left & Top  & Right & Bottom \\
			\midrule
			sorghumHeadyieldTrail & 16   & 618  & 41    & 639    \\
			sorghumHeadyieldTrail & 33   & 1036 & 63    & 1067   \\
			sorghumHeadyieldTrail & 34   & 383  & 75    & 424    \\
			sorghumHeadyieldTrail & 42   & 1019 & 85    & 1059   \\
			sorghumHeadyieldTrail & 43   & 722  & 69    & 745    \\
			sorghumHeadyieldTrail & 44   & 952  & 73    & 976   \\
			\bottomrule
	\end{tabular}
\end{table}

\begin{figure}[!h]
	\centering
	\includegraphics[width=0.8\linewidth]{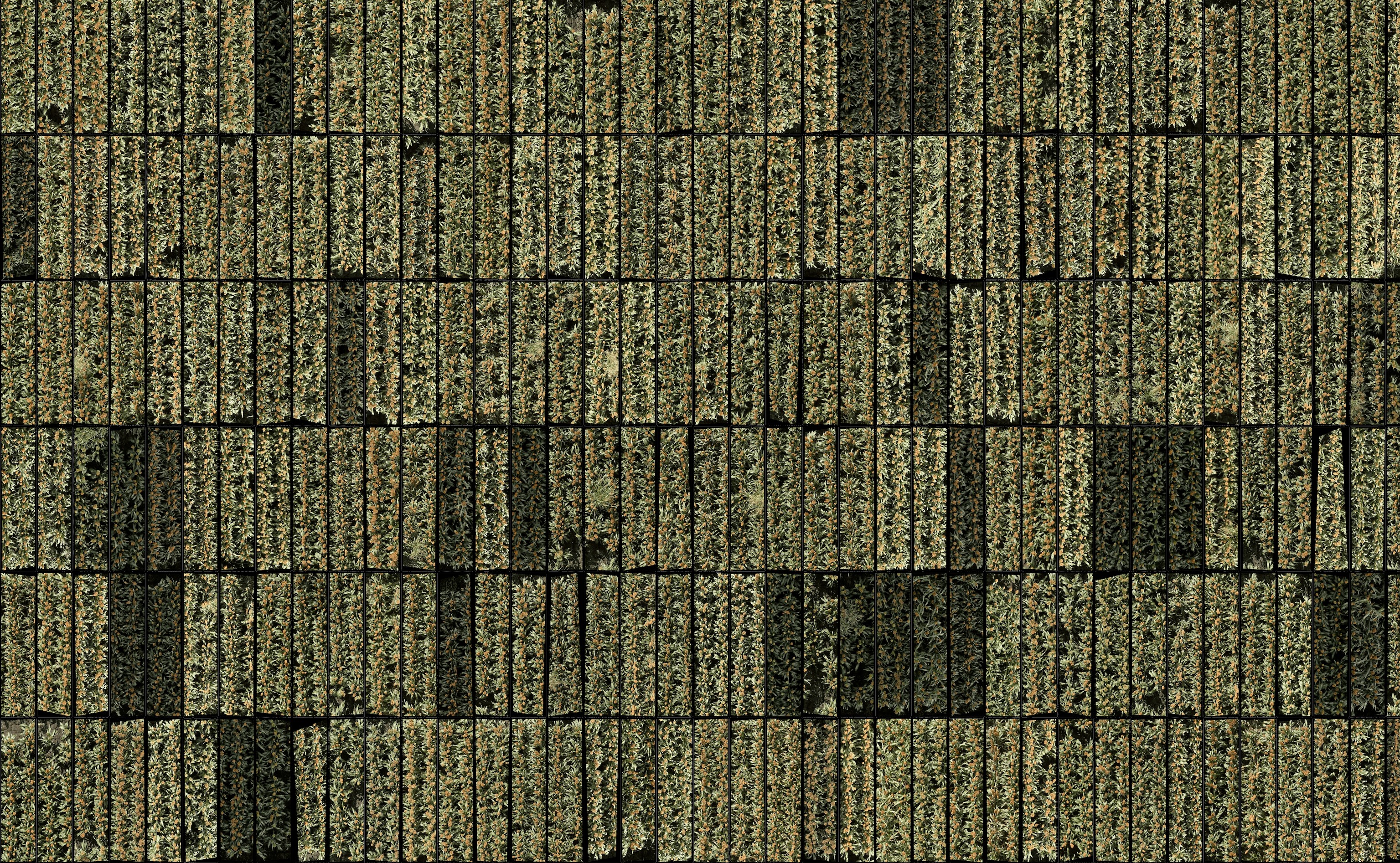}
	\caption{Collage of sorghum images highlighting the variation in photo quality.}
	\label{fig:sorgimagesall}
\end{figure}

To begin the analysis process, we seek to gain an understanding of the uniformity of the imagery data first by visual inspection. At both the image and individual sorghum head levels, we constructed image collages to facilitate within and across image analysis.  For object detection tasks, identification accuracy is made often higher when the images conform to the same uniform lighting, perspective, and dimensional standards across all photos due to increased signal to noise ratios. However, for the images in our training sample, the collage in Figure \ref{fig:sorgimagesall} highlights variation within each of the aforementioned categories.

Inspection shows that within the collection, there are differences in lighting and heights, unexpected fissures within rows, and obstructions blocking the view of the sorghum heads. These effects are more prevalent when scrutinizing the individual heads within an image. In the context of object detection, the non-uniform lighting conditions can be mitigated through a covering of training images. That is, as long as there are a sufficient number of sorghum heads which demonstrate the different lighting conditions, deep learning algorithms will still be able to perform well.

\begin{figure}[!h]
	\centering
	\includegraphics[width=0.35\linewidth]{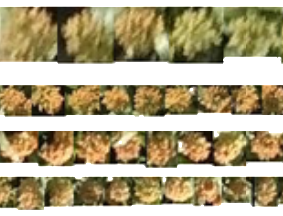}
	\caption{Collage of individual sorghum heads cropped from a random sample of training images.}
	\label{fig:sorgheadcollage}
\end{figure}

When focusing on the sorghum heads themselves, we identified sources of variation in the sorghum heads that can be attributed to factors both endogenous to the plant and exogenous sources from the imaging procedure. At the sorghum head level, variation in the size of the individual sorghum heads, the plant height and the image quality of the individual heads, all point to a lack of uniformity across the images. Outside factors such as drone flight height and variation in landscape height can also artificially increase or decrease the size of the heads within an image, suggesting that some form or control or normalization within our modeling process will be necessary. The sorghum head collage in Figure \ref{fig:sorgheadcollage} showcases examples of image distortions, where identifying singular plant heads is complicated by the existence of low quality image sections which introduce noise inside of the pixel features our future classifier will extract. 

\begin{figure}[h!]
	\centering
	\includegraphics[width=0.5\linewidth]{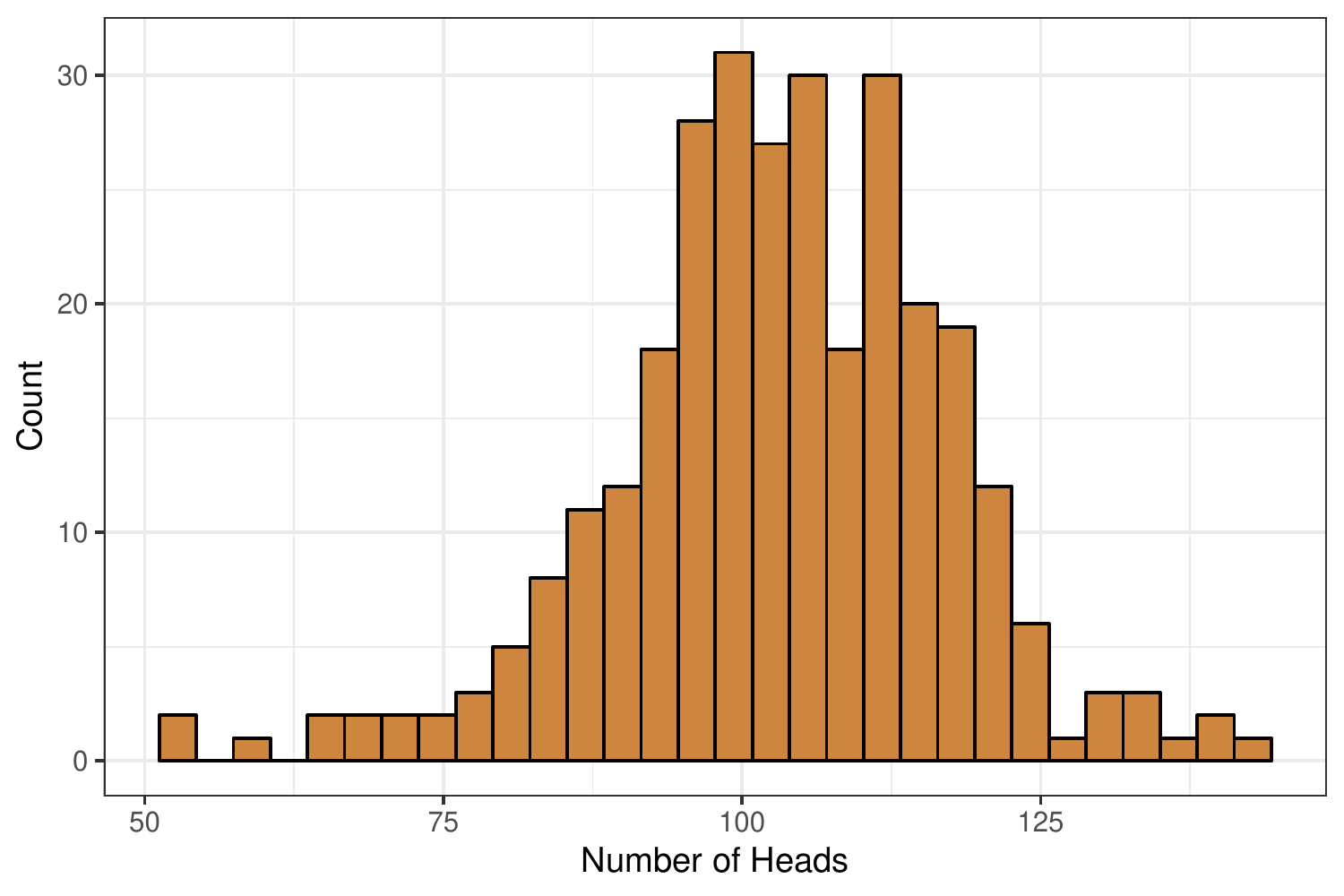}
	\caption{Distribution of the number of sorghum heads within the 300 training images.}
	\label{fig:sorgheaddist}
\end{figure}

To further quantify the quality of the input images, we analyzed the distributions of the number of sorghum heads in an image and the area of the image covered by sorghum. For the collection of 300 training images, there were 30,953 sorghum heads individually labeled with bounding boxes. Collectively, the distribution across all the training samples well approximates a Normal distribution with, 103 sorghum heads contained within each image, on average. This is illustrated in Figure \ref{fig:sorgheaddist}.

\begin{figure}[h!]
	\centering
	\includegraphics[width=.45\linewidth]{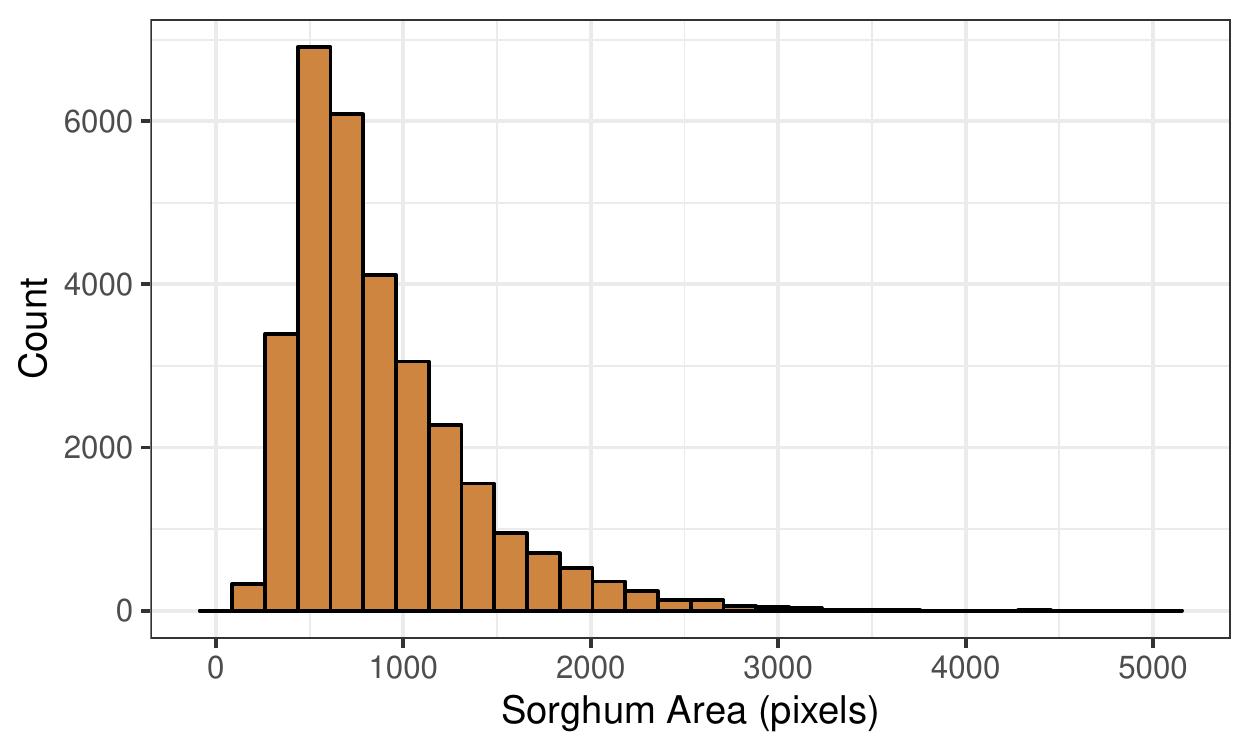}
	\includegraphics[width=.45\linewidth]{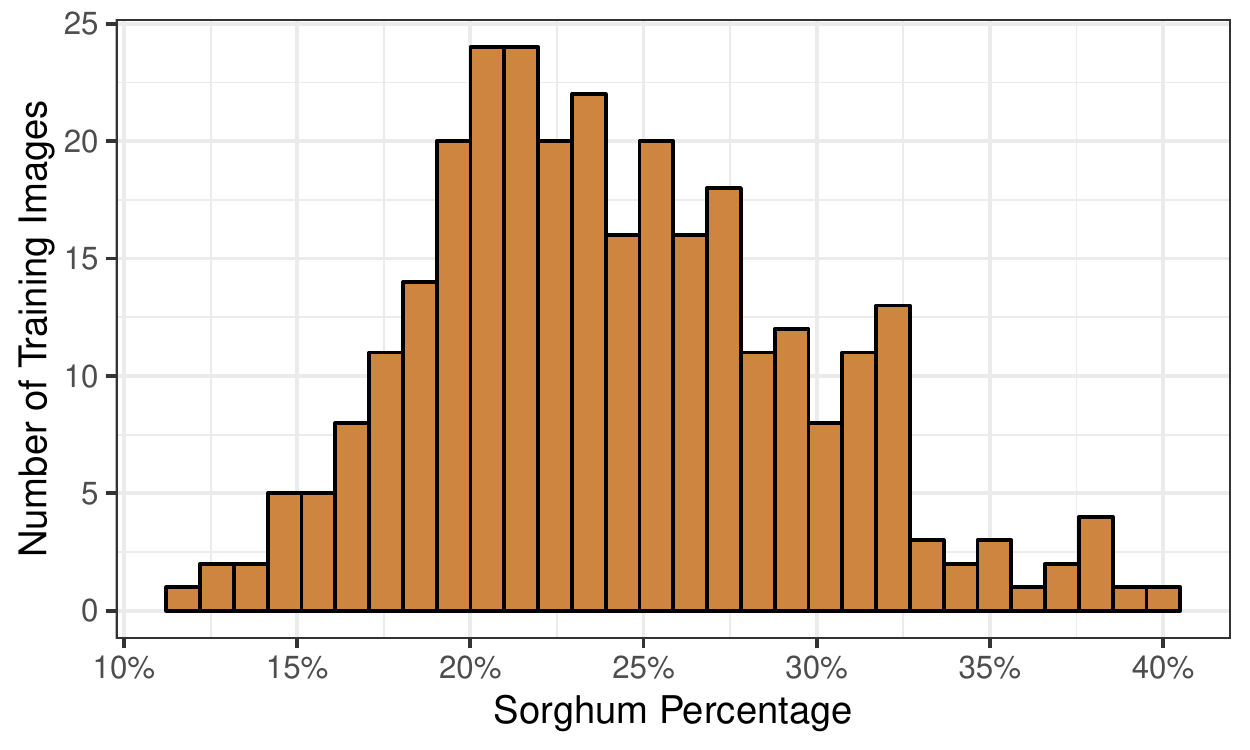}
	\caption{Distribution of sorghum areas and percentages, respectively.}
	\label{fig:sorgarea}
\end{figure}

By taking the total area of the labeled bounding boxes within an image and normalizing by the dimensions, we were able to create an estimate of sorghum head coverage. This estimate was then used to identify outlying images that may have potential data errors.  Shown in Figure \ref{fig:sorgarea}, this metric highlighted a large variation in the sorghum coverage, but inspection of the individual images did not support the removal of any training instances.

These metrics were constructed to deliver guiding intuition for the diagnoses of our modeling process by providing reasonable bounds for model outputs that could inform us whether the results for a training session were acceptable. Results from the model, such as an image with only 3 identified sorghum heads or a sorghum coverage of less than 5\% would immediately identify an error in the model fitting process.

\subsection{Object Counting with YOLO}

Current state of the art methods for object detection and counting leverage deep learning models to perform classification of an image or video (\cite{FMBCAMBBR19,Heinrich}). The specific architectures of these deep learning implementations vary, but all rely on multi-layered neural networks trained with GPUs on exceptionally large image datasets. To perform the sorghum identification and counting task outlined in this paper, the authors leveraged a task tuned version of the YOLO algorithm.  Specifically for this sorghum counting task, we leveraged the pretrained weights from the YOLO's Darknet-53 architecture, a model framework consisting of 53 convolutional layers trained on the ImageNet dataset for the purpose of feature extraction and classification. 

YOLO is a full end-to-end convolutional neural network architecture, constructed to perform object detection through the recasting of the image classification problem as a bounding box regression task. Its approach takes a single image, partitions it into a user specified S $\times$ S square grid, and then attempts to identify objects by encapsulating each grid cell with the most appropriate bounding box and confidence scores for that box. This process is done simultaneously as a feature map of relevant visual constructs is learned by a neural network with a sequential arrangement of convolutional layers, residual blocks, and sampling layers. As a single-shot object detection algorithm, the YOLO model is configured to output an approximate location and the best fit bounding box from labeled images.  During the training process, these images are first partitioned into grids where the deep learning model can then focus on maximizing its ability to select and place anchor boxes, of the appropriate size, over all identifiable classes within each segment simultaneously.  It is this process that not only gives single shot detection algorithms their state-of-the-art speed performance, but also makes their accuracy highly sensitive to the selection of the initializing anchor boxes. 

To perform sorghum identification, using a domain neutral pre-trained YOLO model required modifications to the software implementation. By default, YOLO utilizes a multi-label classifier to identify objects; however, for this task, the configuration file was modified to account for the single sorghum class.  As a result of there being only one class to identify, the intersection over the union for detection was set to 0.70 in order to force the model to return all potential sorghum heads, regardless of the associated class probability. Table \ref{tab:conf} shows the configuration of the Darknet configuration file used for our final results.

\begin{table}[h!]
	\centering
	\caption{Parameters of YOLO Configuration.}	
	\label{tab:conf}
	\begin{tabular}{ll}
		\toprule
		Parameter        & Value  \\
		\midrule                                                                                                                                                              \\
		Mask             & 1, 2, 3, 4, 5                                                                                                                                                       \\
		Anchors          & \begin{tabular}[c]{@{}l@{}}(10,10), (16,16), (19,19), (16,24),\\ (24,20), (23,24), (28,27), (23,35),\\  (32,32), (38,39), (50,50), (60,60), \\ (80,80)\end{tabular} \\
		Classes          & 1                                                                                                                                                                   \\
		Number           & 13                                                                                                                                                                  \\
		Jitter           & 0.3                                                                                                                                                                 \\
		Ignore Threshold & 0.7                                                                                                                                                                 \\
		Truth Threshold  & 1.0                                                                                                                                                                 \\
		Random           & 1.0   \\
		\bottomrule                                                                                                                                                             
	\end{tabular}
\end{table}

\begin{figure}[h!]
	\centering
	\includegraphics[width=0.5\linewidth]{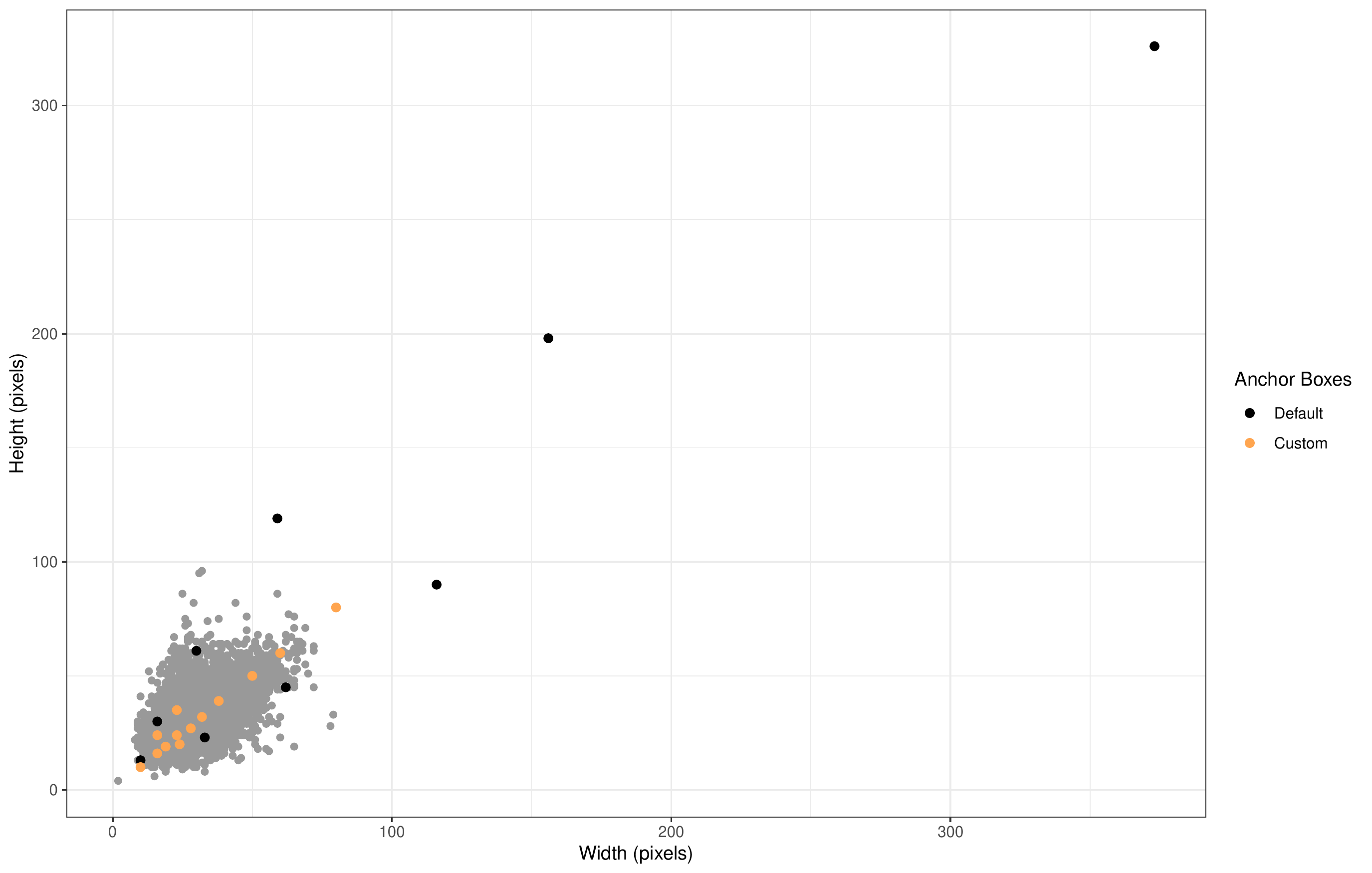}
	\caption{Bounding box dimensions and chosen anchor box sizes in gray and brown, respectively.}
	\label{fig:anchorbox}
\end{figure}

During the weight updating process, at three separate phases, a collection of three anchor box offsets are predicted within each grid. An unconfigured, default YOLO configuration uses a $k$-means clustering method to search for nine candidate anchor boxes from the collection of bounding boxes found in the training set.  The centroids of each group are then converted into rectangular coordinates to form the anchor boxes for which the final objects identified will be bound within. Motivated by the exploratory analysis done previously, we constructed a scatterplot of the height and width of all labeled bounding boxes to visually assess the quality of the algorithmically suggested anchor boxes. Figure \ref{fig:anchorbox}. showcases our findings. The default solution did not propose rational anchor boxes that would sufficiently account for the variation in sorghum head size. The anchor boxes selected via the $k$-means algorithm did not cover segments on the extreme lower tail of the bounding box distribution and, similarly, suggested multiple anchor boxes whose sizes exceeded the largest bounding areas found in the training set by a factor of three.

\begin{figure}[h!]
	\centering
	\includegraphics[width=0.5\linewidth]{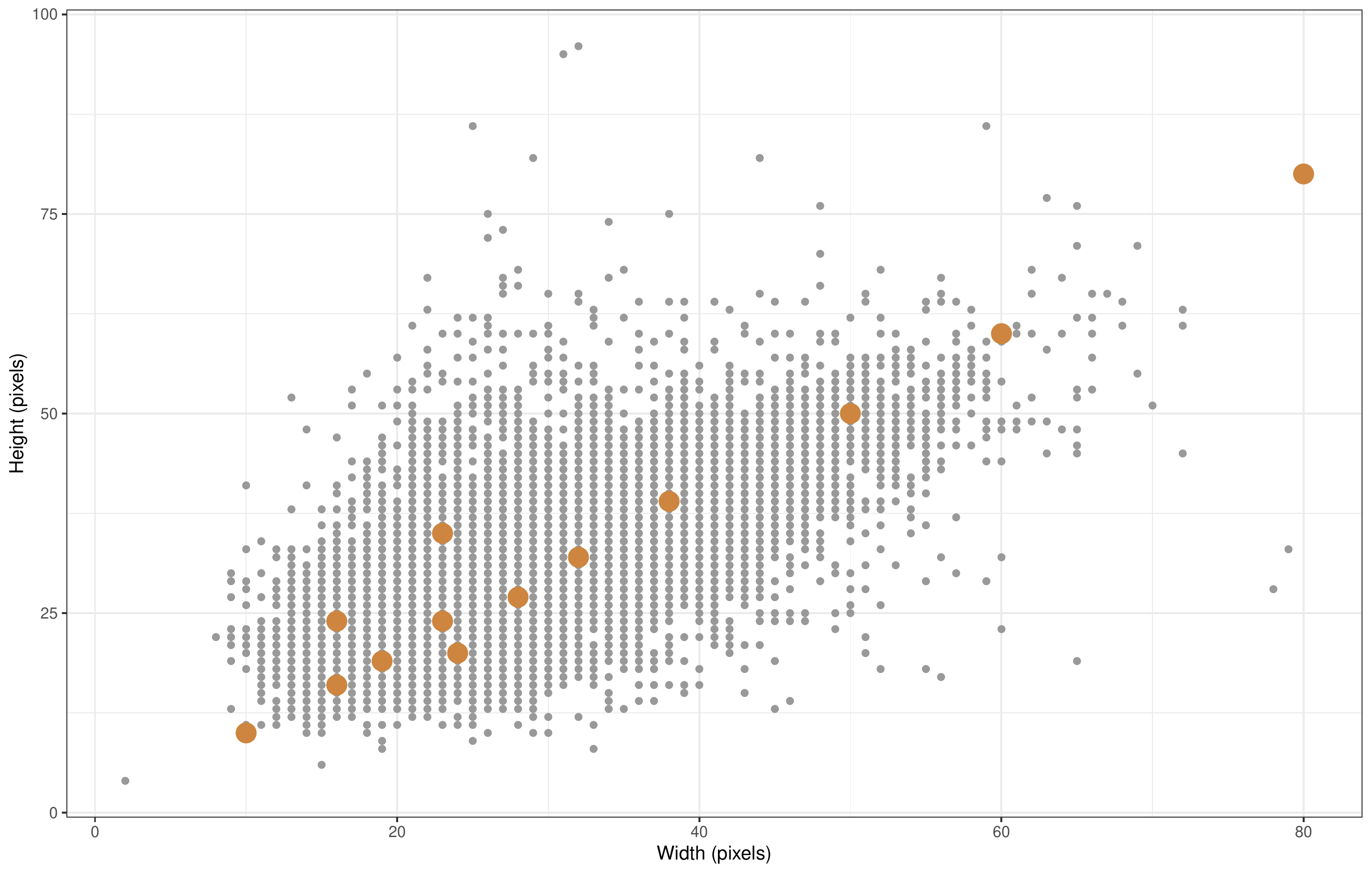}
	\caption{Zoomed-in bounding box dimensions and chosen anchor box sizes in grey and brown, respectively.}
	\label{fig:customanchorbox}
\end{figure}

To overcome this deficiency, we approximated a linear model through the bounding boxes found in our training set and sampled nine anchor boxes along the line. For better coverage of smaller heads at the extreme lower tail of the bounding box distribution, we created a 10 $\times$ 10 anchor box. The remaining anchor boxes were placed in regions of above average variance. Highlighted in brown, the 13 anchor boxes used for prediction can be seen in Figure \ref{fig:customanchorbox}. As this figure demonstrates, we provide significantly better coverage than the default $k$-means algorithm anchor boxes. It is this trick that allows us to be able to detect sorghum heads of various sizes.

The final configuration adjustment to the original architecture consisted of assigning the anchor boxes to various up sampling layers where we included three boxes in the first layer, four in the second layer, and six boxes in the final layer. Lastly, to update the model weights, about 24 hours of additional training was performed on an Ubuntu 18.04.2 LTS workstation with 128 GB of memory, an Intel i7-6850k and a single NVidia GeForce 1080.

\section{Results and Discussion}\label{sec3}

In this section, we provide the summary of our YOLO implementation with parameter-tuned anchor boxes. However, before we begin our discussion on the results, we must first explain how we are measuring the task of detecting sorghum heads. For classification tasks, domain specific measures, which are often dependent on the structure of the data, are required to properly codify the concept of model performance. In object detection problems, evaluation is non-trivial because there are two distinct tasks:

\begin{enumerate}
	\item Determining whether an object exists in the image
	\item Determining the location of the object
\end{enumerate}

We briefly discuss these two measures, but direct the reader to the cited references for more details.
For measuring the existence of objects, a popular metric in measuring the accuracy of object detectors is the average precision (\cite{McFee2010}). Succinctly, average precision is the area under the precision-recall curve which visualizes the true positive rate and the positive predictive value for a predictive model (\cite{Davis:2006:RPR:1143844.1143874}). We then define mean average precision (mAP) as the average precision over all recall values. To measure the location accuracy of our predicted bounding box, we utilize the intersection over union (IoU) which summarizes how well the ground truth object overlaps the object boundary predicted by the model 
(\cite{Rezatofighi_2019_CVPR}). Model object detections are determined to be true or false depending upon the IoU threshold, that is, an object is only detected if the IoU is above a certain level. If IoU = 1, then the ground truth bounding box is equal to the predicting bounding box in both size and locality. Whereas an IoU = 0 would indicate that the intersection of both boxes is empty. Since IoU provides a threshold for object detection, a lower IoU is likely to result in more objects being detected but lower precise localization.

The output of our object detection procedure resulted in separate files for each image with bounding box dimensions and a confidence score for every identified object which is an estimated probability that the object is found within the predicted bounding area. This is shown in Table \ref{tab:test}.  This structure is akin to the one supplied with the training examples.

\begin{table}[h!]
	\centering
	\caption{Sample of identification confidence for sorghum heads in the test set.}
	\label{tab:test}
	 \begin{tabular}{llllll}
			\toprule
			Class Name            & Confidence & Left & Top & Right & Bottom \\
			\midrule
			sorghumHeadyieldTrail & 0.981597   & 26   & 448 & 58    & 477    \\
			sorghumHeadyieldTrail & 0.977624   & 73   & 790 & 104   & 819    \\
			sorghumHeadyieldTrail & 0.975990   & 66   & 132 & 105   & 167    \\
			sorghumHeadyieldTrail & 0.974732   & 231  & 538 & 266   & 568    \\
			sorghumHeadyieldTrail & 0.968000   & 196  & 658 & 225   & 684    \\
			sorghumHeadyieldTrail & 0.965784   & 242  & 627 & 275   & 656   \\
			\bottomrule
	\end{tabular}
\end{table}

\begin{figure}[!h]
	\centering
	\includegraphics[width=0.26\linewidth,angle=90]{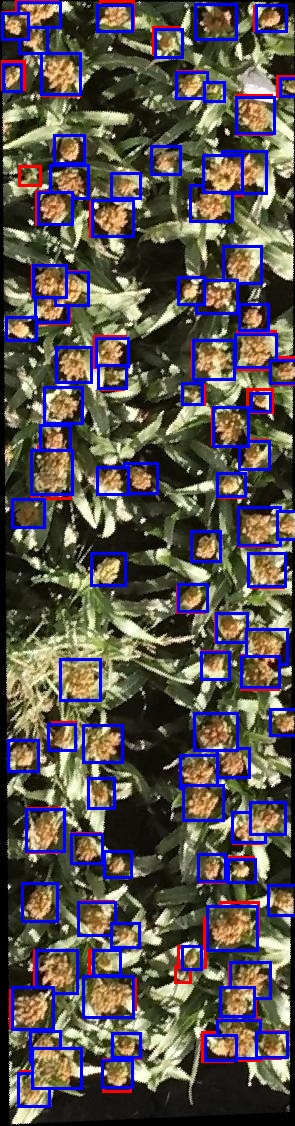}
	
	\vspace{.1cm}
	
	\includegraphics[width=0.26\linewidth,angle=90]{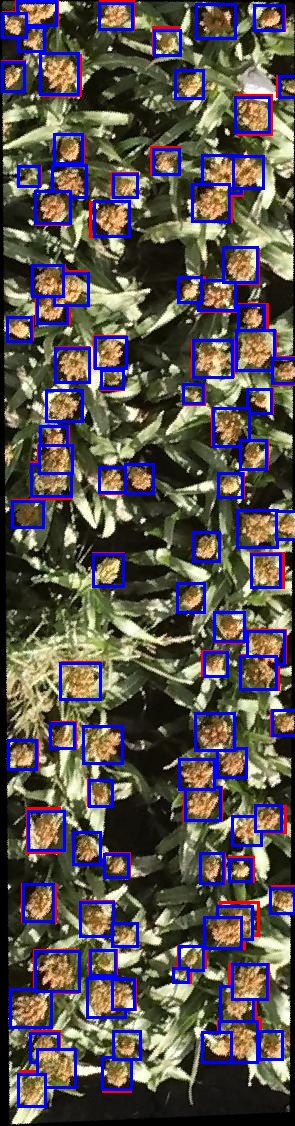}
	\caption{True bounding boxes and predicted bounding boxes; red and blue, respectively of default YOLO (top) and parameter-tuned YOLO (bottom).}
	\label{fig:comparison}
\end{figure}

For the labeled training samples, bounding boxes can be overlaid on the image to show deviations between the true and predicted locations of sorghum heads. This is illustrated in Figure \ref{fig:comparison} in red and blue, respectively.  In the example images, the concepts of detection and intersection over union can be clearly observed. False negatives exist over sorghum heads that have true red labels, but lack a corresponding predicted blue bounding box label. For the sorghum heads found by the prediction procedure, the overlapping intersection between the coordinates give an assessment of the coverage. The incorrect localization of the predicted bounding box and the missed identification of sorghum heads were the dual inhibitors for achieving high mAP scores for our early modeling iterations. Moreover, we notice that the parameter-tuned YOLO model is better able to detect small sorghum heads. This is something that would not be attainable with the default anchor boxes. It is this flexibility in tuning the anchor boxes that allows for superior object localization than otherwise would be possible.

Optimizing over both objectives, by utilizing the anchoring box adjustment technique discussed earlier, resulted in our final training score of 0.99 mAP and the score of 0.95 mAP on the final test set as shown in Table \ref{tab:results}. Since our YOLO model is able to obtain a mAP of 0.99 in the training set, we take that to imply that the parameter-tuned anchor boxes do provide sufficient coverage for the sorghum heads as opposed to the default YOLO model.

\begin{table}[!h]
	\centering
	\caption{Numerical results of default and parameter-tuned YOLO models in mAP.}
	\label{tab:results}
	\begin{tabular}{lll}
		\toprule
		Model                & Training Error & Testing Error \\
		\midrule
		Default YOLO         & 92.0\%         &               \\
		Parameter-tuned YOLO & 99.0\%         & 95.0\%        \\
		\bottomrule
	\end{tabular}
\end{table}

%Table \ref{tab:detectionresults}, we display the accuracy of the detection of sorghum heads as calculated by $R^2$, the coefficient of determination, between the predicted number of sorghum heads and the actual number of sorghum heads in each image. 
%The models which we compare to are that of \cite{Guo2018, Ghosal2019}.
% We notice that in accurately counting the number of sorghum heads, the parameter-tuned YOLO method is superior to the RetinaNet and the quadratic-SVM with less complexity. Although YOLO and RetinaNet are both state-of-the-art object detection methods, we associate the superior performance of YOLO due to the fact that the tuned anchor boxes can account for the smaller sorghum heads that would otherwise be missed by other algorithms.

%\begin{table}[!h]
%	\centering
%	\caption{Detection results in $R^2$ for the sorghum heads detected in the test set versus the actual count.}
%	\label{tab:detectionresults}
%	\begin{tabular}{ll}
%		\toprule
%		Model                & $R^2$     \\
%		\midrule
%		Default YOLO         & 0.90 \\
%		Parameter-tuned YOLO & 0.95 \\
%		%RetinaNet     (\cite{Guo2018})       & 0.89 \\
%		%Quadratic-SVM    (\cite{Ghosal2019})    & 0.88  \\
%		\bottomrule
%	\end{tabular}
%\end{table}

\begin{figure}[h!]
	\centering
	\includegraphics[width=.49\linewidth]{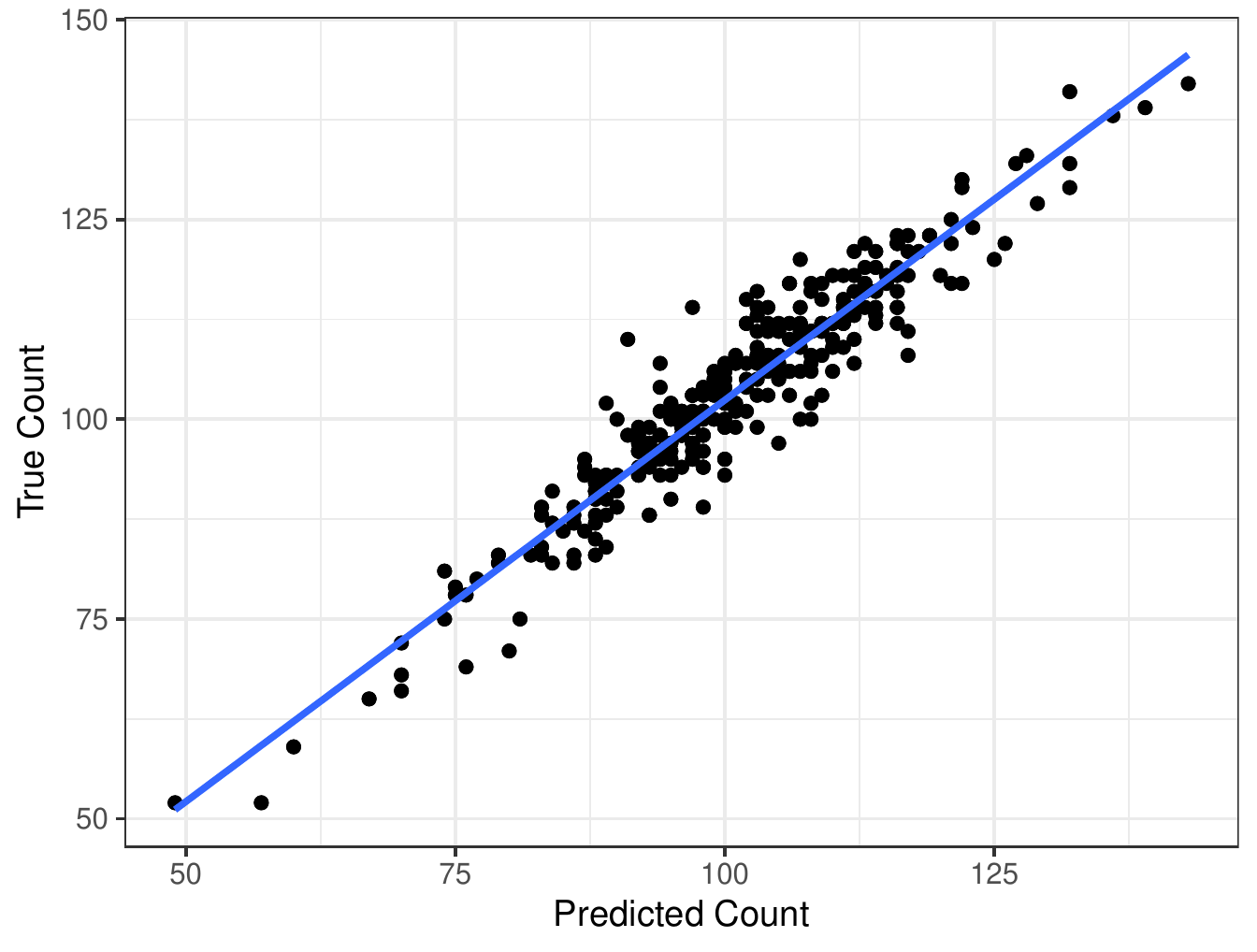}
	\includegraphics[width=.49\linewidth]{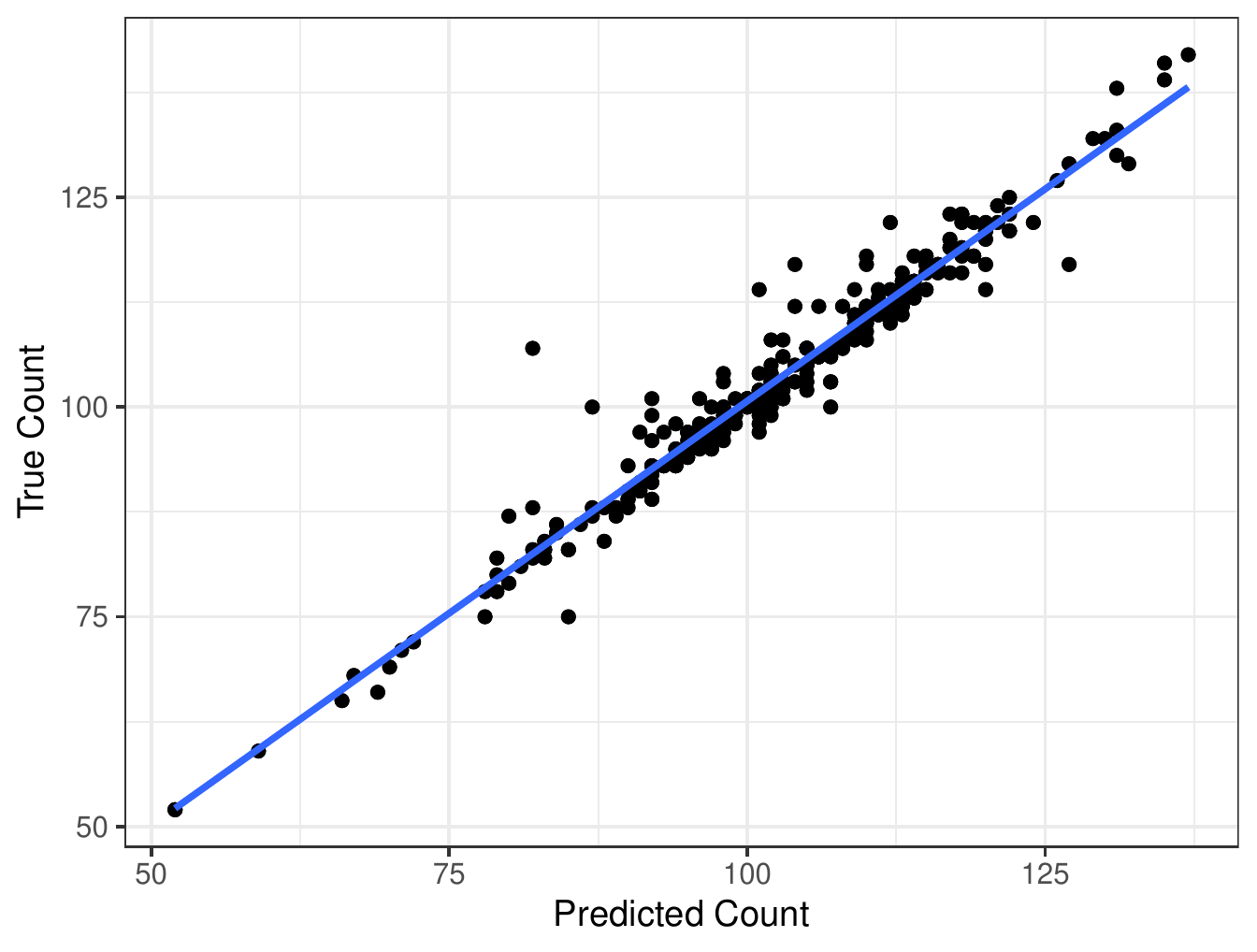}
	\caption{True count versus predicted count of default model and parameter-tuned anchor box model, respectively. The default model has an R2 value of 0.9016 while the parameter-tuned anchor box model has an R2 of 0.9513 (IoU threshold = 0.70 for both).}
	\label{fig:R2}
\end{figure}

From Figure \ref{fig:R2}, it is immediately noticeable that there is indeed a noticeable improvement by tuning anchor boxes. We believe the main performance driver differentiating our parameter-tuned YOLO model from the default lies in the detection of smaller sorghum heads, which would otherwise be excluded by YOLO's embedded anchor box detection algorithm. The R2 value of the parameter-tuned YOLO model displays greater object detection precision than the default YOLO model given an IoU threshold of 0.70.

\section{Conclusion and Future Work}

In this work, we provide a novel example of the use of visual inference to inform the selection of tuning parameters for the improved accuracy of a one shot object detection model. Specifically, we show how exploratory data analysis on the training bounding box sizes can provide a powerful context for assessing the quality of the anchor boxes selected for single shot object detection algorithms. Once tuned, the YOLO model architecture can be used for the identification and counting of phenotypic traits from aerial images. We are able to obtain an mAP of 0.95. 

Previous trials suggest additional performance gains can be found through additional training epochs and further refinement of the anchor box selection. Extensions of this research will involve the pursuit of these alternative approaches. It is also possible to realize additional gains in accuracy by enlarging the solution space search by increasing the anchor boxes used and then training with high class GPUs or TPUs.

To expand to real-time detection of sorghum images and to support other high throughput phenotyping systems, a TinyYOLO implementation may be a computationally efficient enough model to be stored and run on a Raspberry Pi.  This would allow for the automatic detection and counting of sorghum heads directly on the drones in real-time (\cite{DBLP:journals/corr/abs-1811-05588,padala2022optimized}). Moreover, a future research direction could be to utilize the color and size of sorghum heads as an in-season measure of the quality of the plot. This would enable farmers to make real-time decisions without the need to manually inspect their entire field.

\section{Conflict of Interest}

There are no conflicts of interests with regard to this manuscript.

\bibliographystyle{elsarticle-harv}  
\section{References}
\bibliography{Sorghum}

\end{document}